\documentclass[letterpaper, 10 pt, conference]{IEEEtran}

\IEEEoverridecommandlockouts                              

\usepackage{graphicx} 
\usepackage{amsmath} 
\usepackage{amssymb}  
\usepackage{booktabs}
\usepackage{hyperref}
\usepackage[letterpaper, top=54pt, bottom=73pt, left=45pt, right=45pt, nohead, nofoot]{geometry}

\title{\LARGE \bf
Learning Diverse Humanoid Tasks via Synthetic Video Scenarios without Real World Data
}

\author{Yun-Hao Tsai, Cong-Thanh Vu, and Yen-Chen Liu
\thanks{This work was supported in part by the Higher Education Sprout Project, Ministry of Education, to the Headquarters of University Advancement at National Cheng Kung University (NCKU), and by the National Science and Technology Council (NSTC), Taiwan, under Grant NSTC 114-2628-E-006-010.}
\thanks{All authors are with the Department of Mechanical Engineering, National Cheng Kung University (NCKU), Tainan 70101, Taiwan. Email: {\href{mailto:sean901109@gmail.com}{\texttt{sean901109@gmail.com}}, \href{mailto:vuthanh.cdt@gmail.com}{\texttt{vuthanh.cdt@gmail.com}}, \href{mailto:yliu@mail.ncku.edu.tw}{\texttt{yliu@mail.ncku.edu.tw.}}}}
}
\begin{document}

\maketitle
\thispagestyle{empty}
\pagestyle{empty}

\begin{abstract}
The human-like morphology of humanoid robots grants them exceptional potential for agile and versatile motor capabilities, but it also introduces significant challenges in acquiring complex skills. Traditional Learning-from-Demonstrations methods are often constrained by the high cost of collecting real-world data, the difficulty of capturing motion-specific behaviors, and the limited diversity of demonstrations across individuals. Moreover, even for the same task, humans may execute the motion in multiple distinct ways. In this paper, we propose a new framework that leverages the power of Generative AI to convert textual prompts into realistic and diverse sequences of human body movements, enabling the robot to observe multiple variations of how a single task can be performed. These synthetic demonstrations are then used as a training resource, allowing the robot to learn a broad range of task-execution styles without requiring direct human intervention. We evaluate the proposed method across four simulation scenarios. Experimental results show that the robot not only completes the tasks successfully but also demonstrates strong adaptability to complex variations in motion.
\end{abstract}

\section{Introduction}
In recent years, mobile robots have been increasingly deployed across a wide range of domains \cite{10.1007/978-981-16-2094-2_49}, from assisting in daily tasks \cite{11246444, 11093565}. However, most conventional platforms rely on wheeled locomotion, which limits their effectiveness in unstructured or highly dynamic environments. Humanoid robots have therefore become a major focus in modern robotics due to their superior adaptability in human-centered settings. With a morphology modeled after the human body, these robots have the potential to replace humans in labor-intensive tasks and perform complex motor skills \cite{uthai2025opportunities}. Nevertheless, coordinating systems with many degrees of freedom and maintaining stability in a floating-base structure remain difficult challenges, especially when aiming to reproduce natural and efficient human locomotion \cite{gu2025humanoid}.

Passive dynamic walking provides a promising and biologically inspired solution. McGeer’s seminal work \cite{mcgeer1990passive} showed that simple mechanical systems can walk stably down a slope using gravity alone, compensating energy losses from friction and impact without any active control. Subsequent research explored improvements such as optimized foot geometry \cite{4376312}, nonlinear stability analysis \cite{9867274}, and designs that integrate passive mechanics with autonomous locomotion \cite{6027078}. However, purely passive systems suffer from limited disturbance rejection, narrow operating ranges, and strong sensitivity to environmental variations.

To overcome these limitations, recent work has shifted toward data-driven methods. For example, Vu et al. \cite{11203150} use Reinforcement Learning (RL) to refine locomotion patterns rooted in passive dynamics, leveraging the naturally emerging gait properties to achieve more robust and adaptive behaviors.

Although passive locomotion offers high energy efficiency, efficiency alone is not sufficient for humanoid robots intended to operate in everyday human environments. Beyond maintaining balance or walking at a fixed speed, robots must exhibit flexible, natural, and task-aware whole-body motions. In real scenarios, locomotion is tightly coupled with task execution \cite{liao2025beyondmimic}. Robots must not only move between locations but also manipulate objects, navigate confined spaces, carry loads, or adjust posture while interacting with their surroundings \cite{10415857}. Achieving coordinated whole-body motion that balances lower-body stability with upper-body dexterity remains a highly challenging nonlinear control problem \cite{cheng2024expressive}.

To enable more expressive motion, data-driven imitation-learning frameworks such as DeepMimic \cite{10.1145/3197517.3201311} and Adversarial Motion Priors (AMP) \cite{10.1145/3450626.3459670} have demonstrated impressive performance in human motion synthesis. Originally developed for character animation in physics-based simulation, these methods have since been extended to humanoid and quadrupedal robots, with several improvements targeting robustness and sim-to-real transfer \cite{10870445, liao2025beyondmimic, 11128480, zhao2025learning}.
\begin{figure*}[t]
\centering
\includegraphics[scale=0.315,page=1]{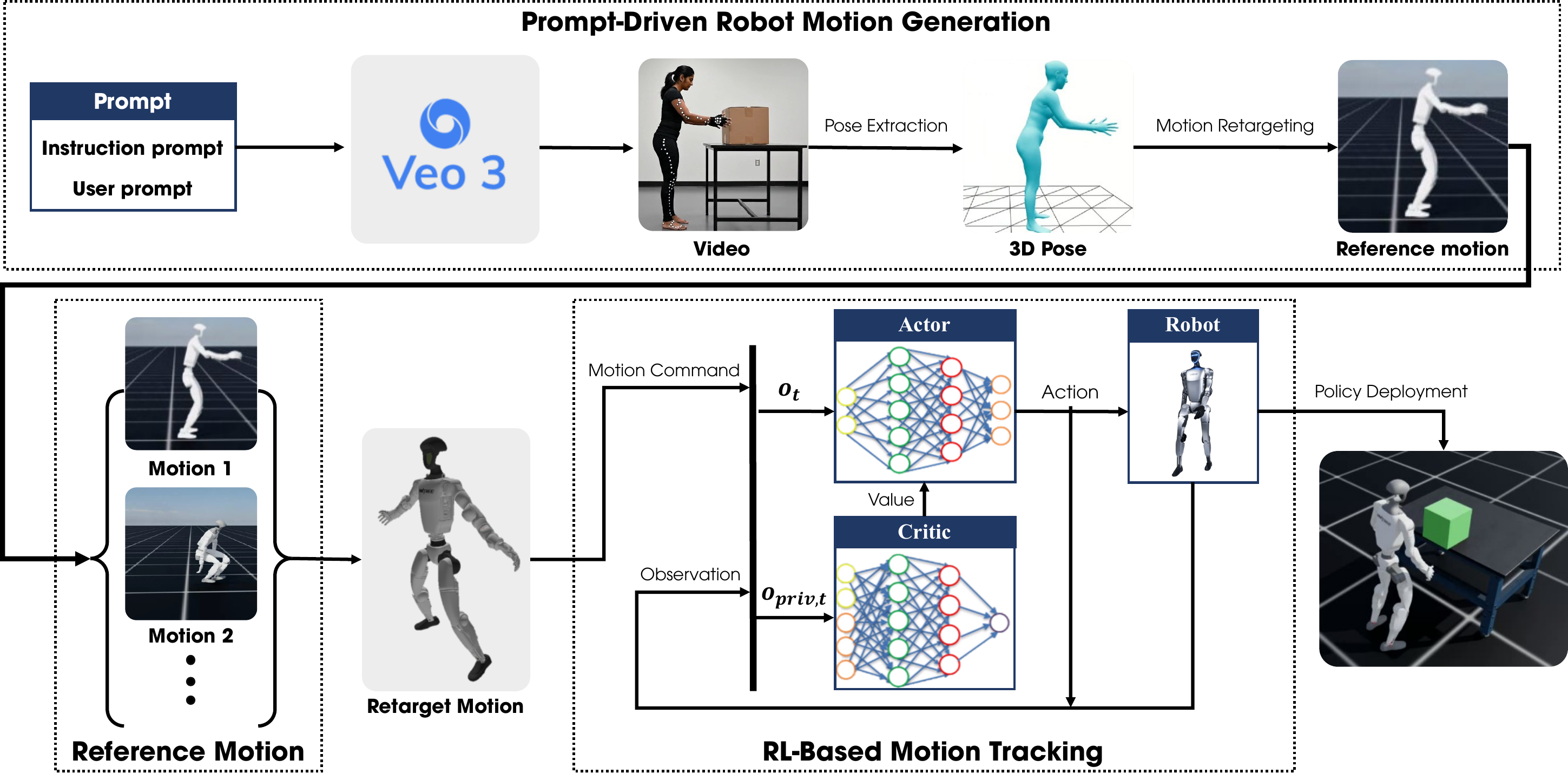}
\caption{Overview of the proposed framework. Multi-modal prompts guide the Veo 3.1 model to generate human motion videos, which are converted into humanoid-compatible reference motions via pose extraction and retargeting. A motion chaining module combines multiple segments into a continuous target motion, which is used as input for RL-based motion tracking.}
\label{fig:Overview}
\end{figure*}

Despite their great potential for generating natural motion and practical applicability, a core challenge remains: these models rely heavily on human demonstration data, especially for complex tasks. Actions such as carrying loads, climbing, maintaining balance while interacting with the environment, or coordinating multiple limbs require highly accurate motion capture systems, tightly controlled recording environments, and meticulous calibration procedures. Moreover, real-world tasks are inherently diverse and highly individualized. Even for the same task, different people perform actions in different ways, producing a wide spectrum of motion with subtle variations rather than following a single pre-recorded demonstration.

Recent advances in Generative AI, including GPT, Gemini, and modern video-generation models, have significantly improved the synthesis and understanding of complex human behaviors. In particular, video-based generative models \cite{kong2024hunyuanvideo,wan2025wan} can now produce realistic and diverse human motions across a wide range of scenarios, creating new opportunities for robot learning, planning, and control. In this work, we present a framework that leverages these generative models to train humanoid robots. By producing realistic human-motion videos from task-specific prompts, the framework enables the creation of diverse action variations for each task, greatly enriching the training data. Our main contributions are:
\begin{itemize}
\item A learning framework that replaces real human demonstrations with synthetic motion data generated through prompt-based generative models.
\item A mechanism for learning tasks from multiple motion variations per prompt, improving behavioral naturalness and diversity.
\item Experimental validation on four scenarios with the Unitree G1, demonstrating robust and versatile motion generation.
\end{itemize}
\section{Architecture Overview}\label{sec:Architecture}
An overview of the framework is shown in Fig.~\ref{fig:Overview}. It consists of three components: (1) converting generated videos into humanoid motion, (2) chaining reference motions, and (3) learning motion policies. Training data are first produced by generative models conditioned on a task-specific prompt, allowing multiple videos to be generated for the same task. Each video is processed to extract human skeletal motion, which is then converted into a robot-compatible representation. To extend policy capability, we introduce a motion-chaining mechanism that merges multiple skills from different videos into a single sequence. The robot is then trained to imitate these references, enabling natural and accurate behavior reproduction. To represent whole-body humanoid motion, we define the robot state as
\begin{equation}
\mathbf{q} = \left[ p_{\text{root}},\, q_{\text{root}},\, \theta_{1}, \ldots, \theta_{n} \right]^T \in \mathbb{R}^{n+7},
\end{equation}
where $p_{\text{root}} \in \mathbb{R}^3$ is the global root position, $q_{\text{root}} \in \mathbb{S}^3$ is the root orientation (unit quaternion), and $\theta_i$ are joint angles.

Although the robot aims to follow a reference trajectory $\mathbf{q}_{\text{ref}}$, strict tracking is impractical due to high DoF and human–robot morphological differences. Enforcing exact alignment of root states often destabilizes balance and degrades task performance. To address this, we adopt an RL framework inspired by DeepMimic~\cite{10.1145/3197517.3201311}, which relaxes strict tracking while encouraging human-like motion. However, prior DeepMimic-style methods rely heavily on MoCap data, which is costly, limited in diversity, and unable to capture the natural variability with which humans perform the same task. Motivated by this limitation, we leverage Generative AI to produce diverse, task-conditioned video demonstrations.
\begin{figure}[t]
\centerline{\includegraphics[scale=0.135,page=1]{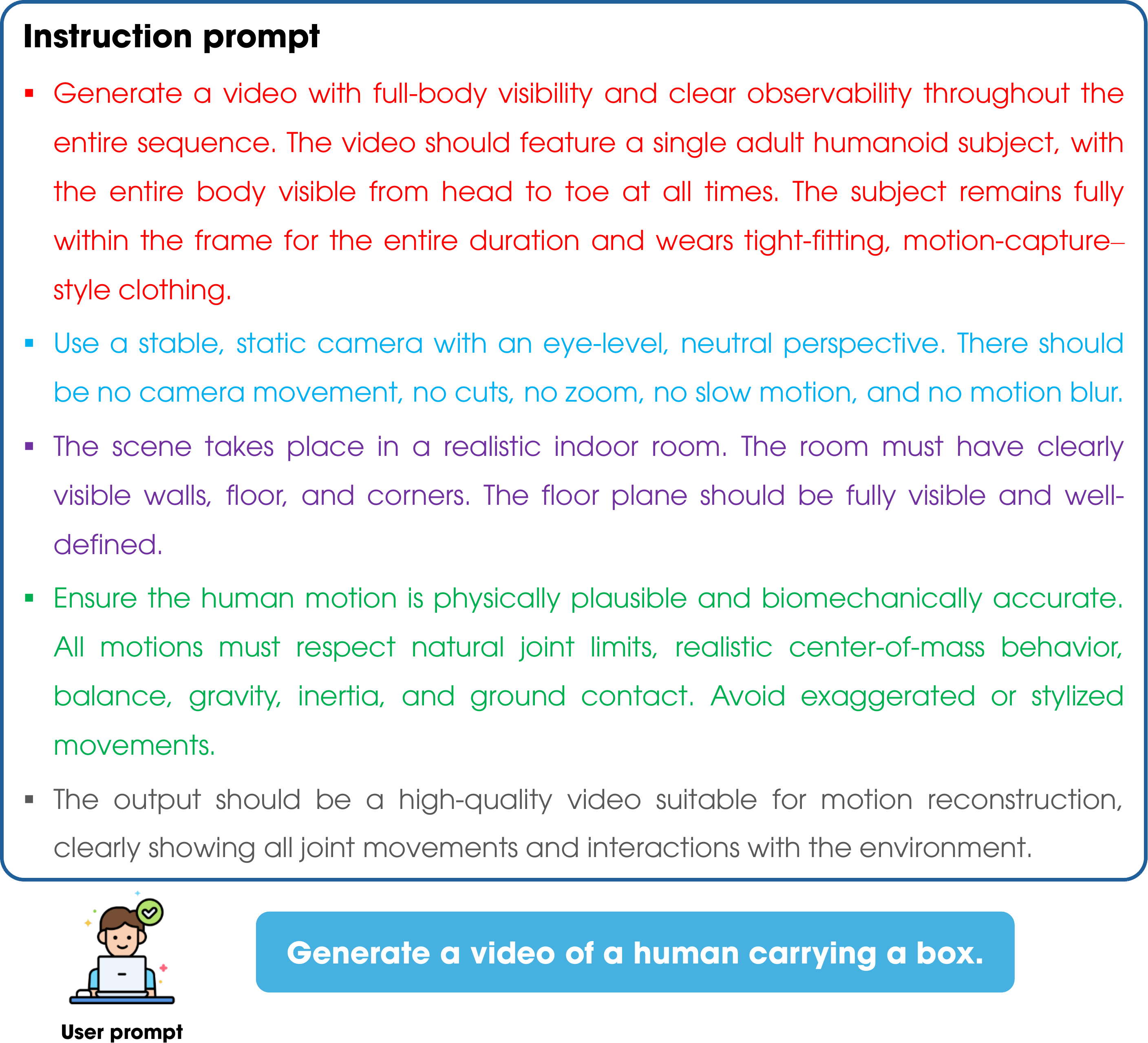}}
\caption{Prompts used in the proposed method: the instruction prompt defines the requirements for generating the video, while the user prompt specifies the desired content of the video provided by the user.}
\label{fig:prompt}
\end{figure}
\section{Prompt-Driven Robot Motion Generation}\label{sec:Motion}
\subsection{Generate Video to Humanoid Reference Motion}
We adopt a motion-conversion pipeline that transforms AI-generated videos into humanoid reference trajectories. A structured prompt is used to ensure consistent visual conditions—full-body visibility, static camera placement, and physically realistic human motion—so that the resulting videos are suitable for downstream reconstruction. For each task, we append a user-specified action prompt to generate videos depicting the desired motion, as shown in Fig.~\ref{fig:prompt}. These constraints improve consistency, physical plausibility, and the overall quality of the extracted data, allowing multiple motion variations to be generated for the same task.

Once the task-specific videos are obtained, processing proceeds in two stages. First, 3D human motion is reconstructed by estimating high-fidelity SMPL-X \cite{pavlakos2019expressive} parameters using a vision-transformer model enhanced with bounding-box cues, body-shape priors, and a temporal module that aggregates information across frames to reduce ambiguity and improve temporal coherence.

\begin{figure}[t]
\centerline{\includegraphics[width=0.36\textwidth,page=1,page=1]{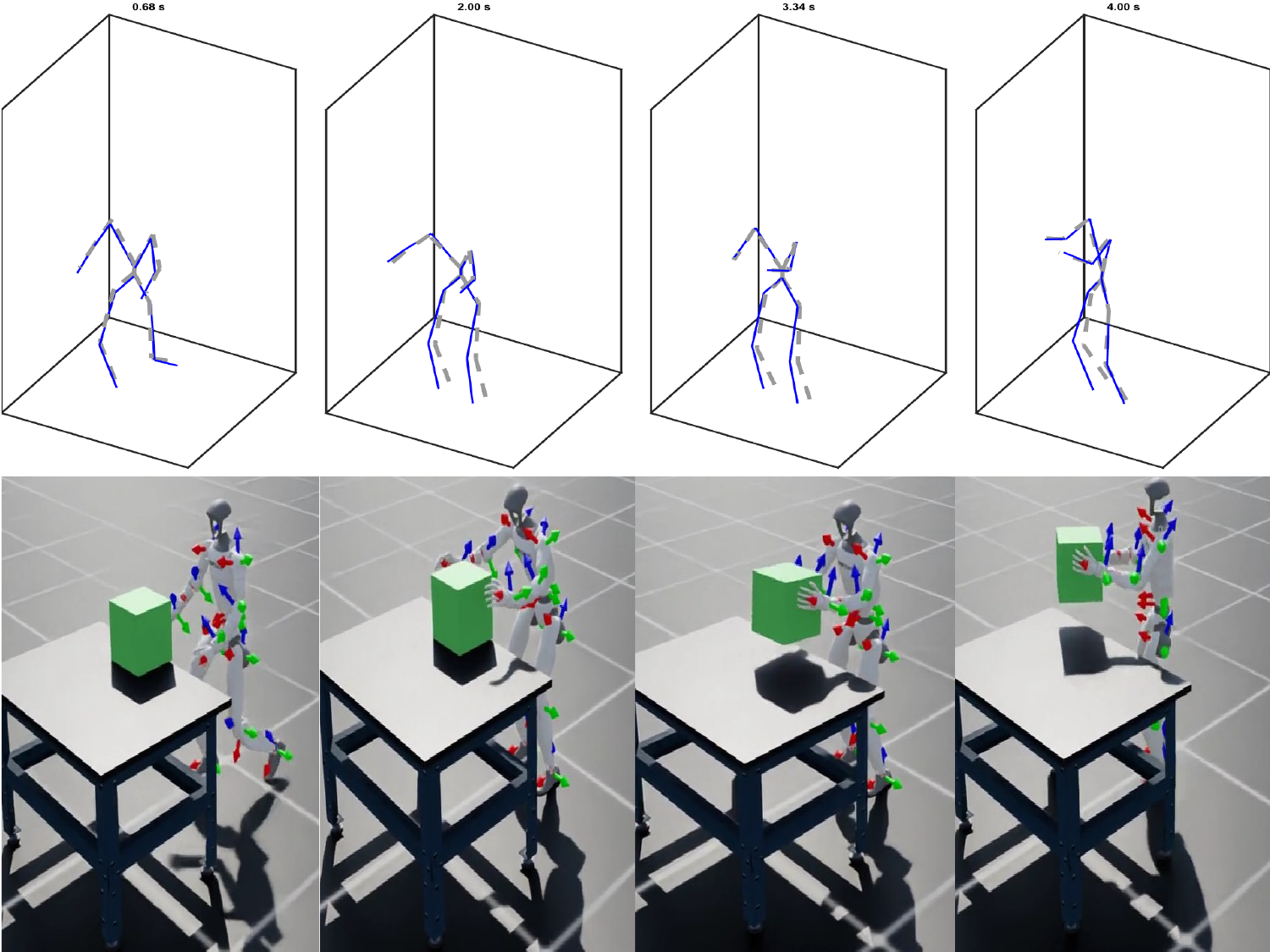}}
\caption{Human skeletal motion is extracted from Generative AI video data and subsequently retargeted to produce robot-compatible motion trajectories.}
\label{fig:Anchor}
\end{figure}

Second, the reconstructed motion is retargeted to the humanoid using GMR \cite{araujo2025retargeting}. The human and robot initial poses are aligned, limb lengths are scaled to compensate for geometric mismatch, and artifacts such as foot sliding and ground penetration are mitigated. A final inverse-kinematics optimization minimizes joint-rotation and end-effector errors under strict joint-limit constraints, producing robot-feasible motion trajectories that preserve the essential characteristics of the original human movements, as illustrated in Fig.~\ref{fig:Anchor}.
\subsection{Motion Stitching}
Because each motion sequence in our framework is extracted from an independently generated video, every sequence is defined in its own local coordinate system. Directly concatenating them leads to physical discontinuities, most notably abrupt changes in the root pose due to differing camera viewpoints and initial human poses, and sudden joint-angle jumps caused by mismatched joint configurations. These discontinuities produce unrealistic transitions and can destabilize control in simulation.

To resolve this, we introduce an automated motion-stitching procedure consisting of two steps. First, root coordinate alignment transforms the second sequence so that its initial root pose matches the terminal root pose of the first sequence, ensuring a continuous and physically consistent root trajectory. Second, joint-configuration smoothing addresses residual discontinuities by inserting a short transition buffer in which joint angles are interpolated smoothly, yielding gradual, physically plausible transitions.

The final stitched sequence is formed by concatenating the first motion, the interpolated transition buffer, and the aligned second motion, producing a continuous and simulation-stable trajectory suitable for downstream policy training.
\section{RL-Based Motion Tracking}\label{sec:rl}
\subsection{Observation Space}
The observation vector follows a root-relative state representation in the robot's local frame.  
All kinematic features are expressed relative to the root to provide invariance to global translations and rotations.  
At each time step $t$, the observation captures the robot's full-body state together with task-specific guidance from the reference motion.  
Specifically, it includes the relative positions and orientations of all links, their linear and angular velocities, the previous action $a_{t-1}$, and the per-link tracking error with respect to the reference motion.  

Formally, the observation vector is defined as:
\begin{equation}
o_t = \left[ 
p^{rel}_{1:n},\; 
q^{rel}_{1:n},\; 
v^{rel}_{1:n},\; 
\omega^{rel}_{1:n},\; 
e_{1:n},\; 
a_{t-1} 
\right],
\end{equation}
where $p^{rel}_{1:n}$, $q^{rel}_{1:n}$, $v^{rel}_{1:n}$, and $\omega^{rel}_{1:n}$ are the link-wise positions, orientations (quaternions), linear velocities, and angular velocities, respectively, all expressed in the root frame. The per-link tracking error is defined as:
\begin{equation}
e_{1:n} = \left( p_{1:n}^{rel} - p_{1:n}^{ref},\; q_{1:n}^{rel} \ominus q_{1:n}^{ref} \right),
\end{equation}
where $p_i^{ref}$ and $q_i^{ref}$ are extracted from the reference motion $\mathbf{q}_{\text{ref}}$.

For the critic network, a privileged observation $o_{\text{priv},t}$ is provided, containing full, noise-free kinematics along with the target reference poses:
\begin{equation}
o_{\text{priv},t} = \left[ 
p^{rel}_{1:n},\; 
q^{rel}_{1:n},\; 
v^{rel}_{1:n},\; 
\omega^{rel}_{1:n},\; 
p^{ref}_{1:n},\; 
q^{ref}_{1:n},\; 
e_{1:n} 
\right].
\end{equation}
Here, $p^{ref}_{1:n}$ and $q^{ref}_{1:n}$ denote the reference link positions and orientations. The actor outputs target joint orientations for all actuated joints, which are subsequently tracked via PD control at the lower-level control stage.
\subsection{Reward Function}
The reward function is composed of two primary components, adapted from DeepMimic \cite{10.1145/3197517.3201311}: the motion tracking reward (imitation objective) and the base reward (regularization penalties).

The motion tracking reward $r^I_t$ measures how accurately the robot follows the reference trajectory, which is encoded in the motion command. This objective is formulated as a weighted sum of terms based on kinematic differences between the robot's state and the reference motion. It is further decomposed into components that reward the matching of joint orientations, joint velocities, end-effector positions, and the center of mass.
\begin{equation}
r^I_t = w_p r^p_t + w_v r^v_t + w_e r^e_t + w_c r^c_t,
\end{equation}
where $w_p$, $w_v$, $w_e$, and $w_c$ are empirically chosen weights.  

Each individual reward component is defined using the root-relative features introduced in the observation space. Specifically, $r^p_t$ corresponds to per-link position tracking, $r^v_t$ to linear and angular velocity tracking, $r^e_t$ to end-effector pose tracking, and $r^c_t$ to center-of-mass tracking.

The pose reward $r^p_t$ encourages the robot to match the joint orientations of the reference motion.  
Based on the defined relative orientations, it is computed as follows, where $\ominus$ denotes the quaternion difference and $\|q\|$ computes the scalar rotation of a quaternion about its axis in radians:
\begin{equation}
r^p_t = \exp \Bigg( -\alpha_p \sum_{i=1}^n \| q_i^{rel} \ominus q_i^{ref} \|^2 \Bigg),
\end{equation}
where $\alpha_p$ is a weighting factor that controls the influence of joint orientation errors on the total reward.

The velocity reward $r^v_t$ is computed from the difference in local joint angular velocities.  
The target angular velocity $\omega_i^{ref}$ is derived from the reference motion:
\begin{equation}
r^v_t = \exp \Bigg( -\alpha_v \sum_{i=1}^n \| \omega_i^{rel} - \omega_i^{ref} \|^2 \Bigg),
\end{equation}
where $\alpha_v$ adjusts the contribution of velocity matching to the overall reward.

The end-effector reward $r^e_t$ encourages the robot’s end-effectors (e.g., hands and feet) to match their corresponding positions in the reference motion.  
Here, $E$ denotes the set of end-effector indices:
\begin{equation}
r^e_t = \exp \Bigg( -\alpha_e \sum_{e \in E} \| p_e^{rel} - p_e^{ref} \|^2 \Bigg),
\end{equation}
where $\alpha_e$ controls the relative importance of end-effector position tracking.

The center of mass reward $r^c_t$ penalizes deviations of the robot’s center of mass $p_c^{rel}$ from the reference center of mass $p_c^{ref}$:
\begin{equation}
r^c_t = \exp \Big( -\alpha_c \| p_c^{rel} - p_c^{ref} \|^2 \Big),
\end{equation}
where $\alpha_c$ weights the influence of center-of-mass deviations on the overall reward.

The base reward is designed to enforce physically plausible and smooth behavior.  
It includes penalties that constrain joint positions within soft limits, reduce high-frequency variations in control actions, and prevent excessive self-collisions.  
Specifically, these terms correspond to the joint limit penalty $r_{\text{limit}}$, the action smoothness penalty $r_{\text{smooth}}$, and the self-collision penalty $r_{\text{contact}}$, which are defined as follows.

The joint limit penalty $r_{\text{limit}}$ discourages the robot from exploring joint configurations that approach mechanical limits.  
It is formulated as a quadratic penalty when a joint angle $q_i$ exceeds the predefined lower bound $q_i^{\min}$ or upper bound $q_i^{\max}$:
\begin{equation}
r_{\text{limit}} = w_{\text{limit}} \sum_{i=1}^n \Big( \max(0, q_i - q_i^{\max})^2 + \max(0, q_i^{\min} - q_i)^2 \Big).
\end{equation}

The action smoothness penalty $r_{\text{smooth}}$ minimizes high-frequency oscillations and ensures continuous, natural control signals.  
It penalizes large variations between consecutive actions output by the policy:
\begin{equation}
r_{\text{smooth}} = w_{\text{smooth}} \| a_t - a_{t-1} \|^2.
\end{equation}

The self-collision penalty $r_{\text{contact}}$ prevents physically impossible self-intersections or undesirable impacts between the robot's own body segments.  
It is computed by penalizing the number of invalid contact events:
\begin{equation}
r_{\text{contact}} = w_{\text{contact}} \sum_{k \in C_{\text{invalid}}} \mathbb{I}_{\text{contact}}(k),
\end{equation}
where $C_{\text{invalid}}$ denotes the set of body links explicitly forbidden from colliding (e.g., arms intersecting the torso), and the indicator function $\mathbb{I}_{\text{contact}}(k)$ equals $1$ if an invalid contact occurs at link $k$, and $0$ otherwise.
\begin{figure*}[t]
\centering
\includegraphics[scale=0.33,page=1]{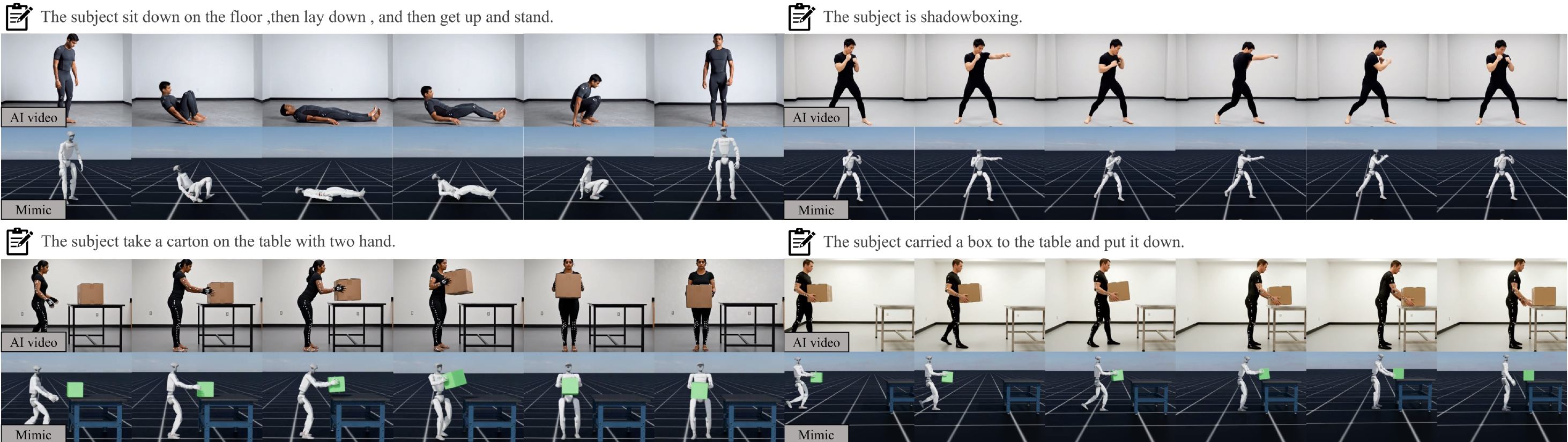}
\caption{Qualitative results from four experiments evaluating policies trained using generative AI with a single prompt. All motions are executed on the humanoid robot via RL-based tracking and include full-body transitions and manipulation tasks such as lie-and-stand, boxing, and pick-and-place.}
\label{fig:motion}
\end{figure*}
\begin{figure*}[t]
\centering
\includegraphics[width=0.85\textwidth,page=1]{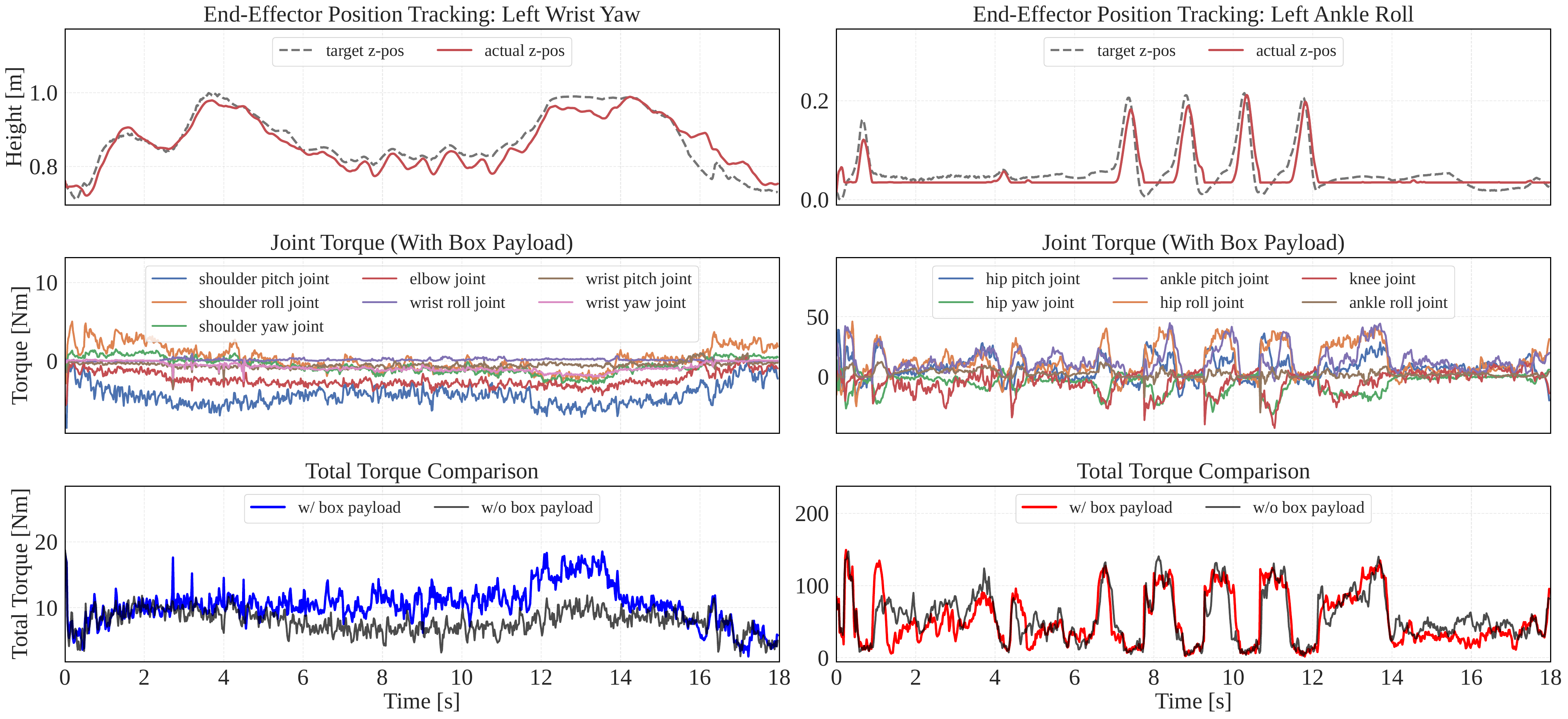}
\caption{Evaluation of the learned policy for the box pick-and-place task, showing upper- and lower-body joint positions, vertical trajectories of the left wrist and ankle, and joint torque profiles with and without a 0.5 kg payload.}
\label{fig:result}
\end{figure*}

By integrating these tracking objectives and regularization penalties, the robot is encouraged to accurately track the reference motion while maintaining stability, smoothness, and physical feasibility. The overall reward function $r_t$ at each time step is defined as:
\begin{equation}
r_t = r^I_t - r_{\text{limit}} - r_{\text{smooth}} - r_{\text{contact}}
\end{equation}

To balance robustness and stability across diverse motions, we apply a minimal set of randomizations targeting key factors affecting generalization: ground friction and restitution, nominal joint configuration (to model joint offset errors in both actions and observations), and torso center of mass. In addition, random velocity perturbations are introduced to enhance robustness against external disturbances. In our training process, the empirical weightings and scaling factors are set as follows. For the imitation objective, the component weights are $w_p = 0.65$, $w_v = 0.1$, $w_e = 0.15$, and $w_c = 0.1$. The corresponding scaling factors for the exponential functions are chosen as $\alpha_p = 2.0$, $\alpha_v = 0.1$, $\alpha_e = 40.0$, and $\alpha_c = 10.0$. For the regularization penalties, the weights are set to $w_{\text{limit}} = 1.0$, $w_{\text{smooth}} = 0.1$, and $w_{\text{contact}} = 0.1$.
\section{Results and Discussion}\label{sec:Results}
\subsection{Experimental Setup}
The proposed framework is trained in the NVIDIA Isaac Lab environment, where CUDA-accelerated parallelism enables efficient large-scale learning. The Unitree G1 policy is trained with 4096 parallel agents on a workstation equipped with an Intel Xeon W5-3435X CPU and an RTX 4000 Ada GPU. Training follows the PPO algorithm \cite{schulman2017proximal} with an asymmetric actor–critic architecture. Both networks are fully connected with three hidden layers of sizes [512, 256, 128] and ELU activations. The policy is supervised using reference trajectories extracted from AI-generated human motion, which provide structured guidance for learning natural and accurate behaviors. We evaluate the method through two studies: (1) assessing the ability of Generative AI to produce videos that yield robot-compatible motion trajectories, and (2) evaluating the performance of the trained policy using these AI-derived reference motions.
\subsection{Motion Generation Evaluation Results}
To evaluate the capability of the generative model in producing diverse and task-consistent motion videos from user-specified prompts, we utilize the Google Veo 3 API. A total of 50 everyday task prompts were tested, and for each prompt, 10 distinct videos were generated depicting the corresponding behavior required for the robot.  

During validation, a task was considered successful if the generated motions were physically plausible and correctly represented the intended behavior. The results indicate that, for most everyday tasks, the generative model produces accurate and diverse motion sequences. However, for high-speed actions such as backflips, certain segments exhibited temporal inconsistencies, leading to unrealistic motion. This highlights the current limitations of the model in generating rapid or highly dynamic movements.  
\subsection{Policy Evaluation Results}
\begin{figure}[t]
\centerline{\includegraphics[scale=0.105,page=1]{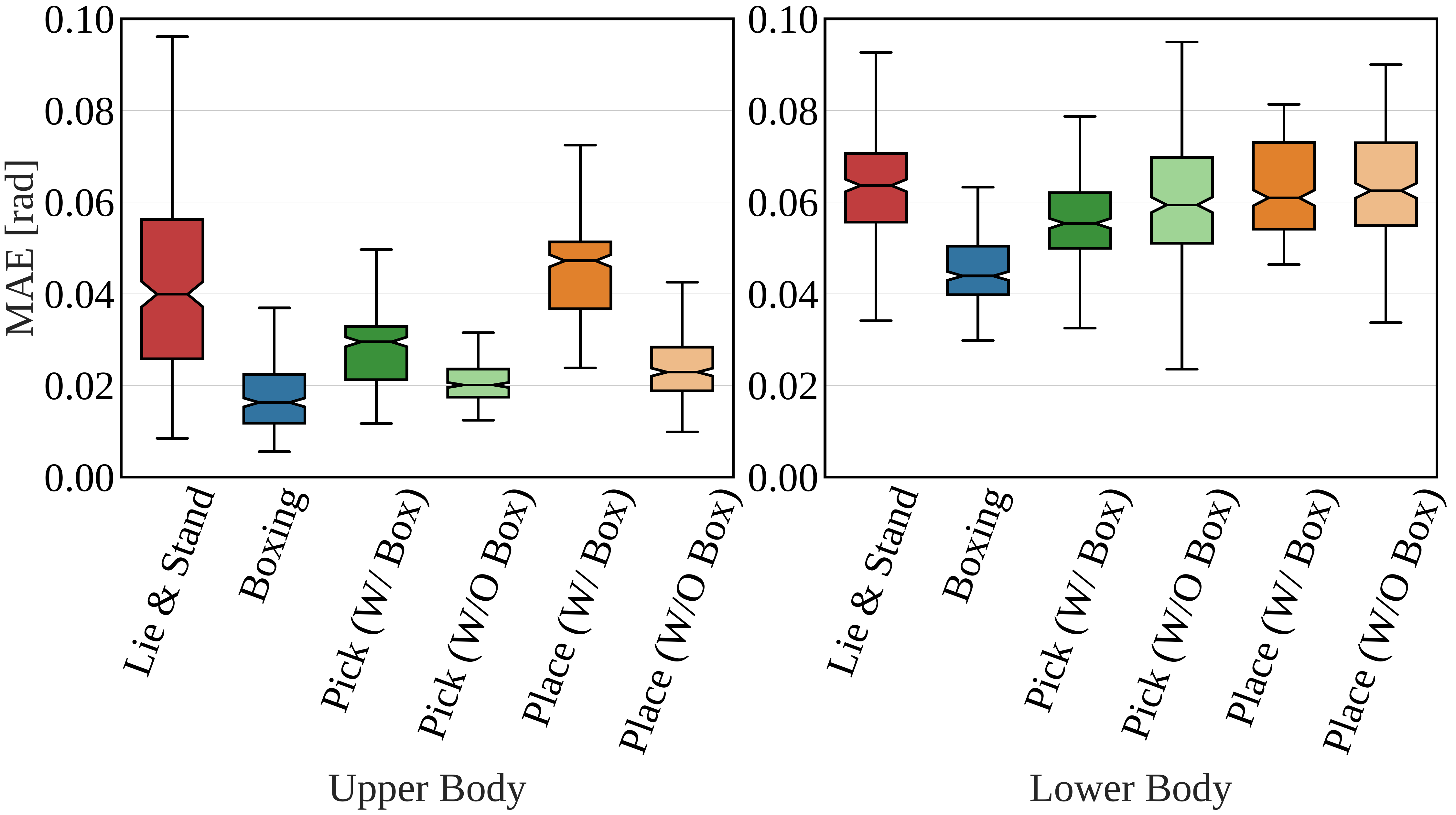}}
\caption{MAE of joint positions for both upper and lower body across the tasks lie-and-stand, boxing, and pick-and-place under loaded and unloaded conditions.}
\label{fig:MAE}
\end{figure}
Fig.~\ref{fig:motion} illustrates action sequences generated from AI videos and their corresponding policy executions. Using user-specified prompts, the proposed method generates diverse motion variations for the same task, producing physically feasible trajectories while preserving the structure of human motion. The robot can then follow these reference motions to imitate target behaviors in a natural manner.

Fig.~\ref{fig:result} presents a detailed evaluation of the pick-and-place task. The close alignment between target and executed trajectories along the vertical axis of the wrist and ankle indicates accurate spatial tracking. The joint torque profiles show coordinated upper- and lower-body control for stable manipulation. Under a 0.5 kg payload, the lower body remains largely unchanged, while the upper body exhibits increased torque to compensate for the load.

Quantitatively, the mean absolute error (MAE) of joint positions across the body ranges from 0.04 m to 0.07 m for dynamic tasks such as boxing and pick-and-place. Under loaded conditions, upper-body errors increase noticeably, while lower-body errors remain stable, indicating that external loads primarily affect upper-body control.

Overall, the results demonstrate that generative video data can provide diverse and realistic motion for robot training, reducing reliance on real demonstrations while improving behavioral diversity and robustness.
\section{Conclusions}\label{sec:Conclusion}
In this study, we present a framework that leverages Generative AI to synthesize human motion data and train humanoid robots without relying on traditional motion capture systems. By converting prompt-driven video generation into humanoid-compatible reference motions, the proposed pipeline enables scalable and diverse motion acquisition for complex tasks. In addition, the reference motion chaining mechanism allows multiple motion segments to be seamlessly integrated, enabling a single policy to execute extended and compound behaviors. Results in a simulation environment demonstrate that the proposed method can successfully learn a wide range of dynamic and manipulation tasks. The learned policies achieve accurate motion tracking while maintaining stability using data generated by the generative model. Future work will focus on improving sim-to-real transfer, enhancing physical consistency in generated motions, and extending the framework to more complex tasks.
\bibliographystyle{ieeetr}
\bibliography{references}

@article{araujo2025retargeting,
  title={Retargeting matters: General motion retargeting for humanoid motion tracking},
  author={Araujo, Joao Pedro and Ze, Yanjie and Xu, Pei and Wu, Jiajun and Liu, C Karen},
  journal={arXiv preprint arXiv:2510.02252},
  year={2025}
}

@article{liao2025beyondmimic,
  title={Beyondmimic: From motion tracking to versatile humanoid control via guided diffusion},
  author={Liao, Qiayuan and Truong, Takara E and Huang, Xiaoyu and Gao, Yuman and Tevet, Guy and Sreenath, Koushil and Liu, C Karen},
  journal={arXiv preprint arXiv:2508.08241},
  year={2025}
}

@article{kong2024hunyuanvideo,
  title={Hunyuanvideo: A systematic framework for large video generative models},
  author={Kong, Weijie and Tian, Qi and Zhang, Zijian and Min, Rox and Dai, Zuozhuo and Zhou, Jin and Xiong, Jiangfeng and Li, Xin and Wu, Bo and Zhang, Jianwei and others},
  journal={arXiv preprint arXiv:2412.03603},
  year={2024}
}

@article{wan2025wan,
  title={Wan: Open and advanced large-scale video generative models},
  author={Wan, Team and Wang, Ang and Ai, Baole and Wen, Bin and Mao, Chaojie and Xie, Chen-Wei and Chen, Di and Yu, Feiwu and Zhao, Haiming and Yang, Jianxiao and others},
  journal={arXiv preprint arXiv:2503.20314},
  year={2025}
}

@inproceedings{pavlakos2019expressive,
  title={Expressive body capture: 3d hands, face, and body from a single image},
  author={Pavlakos, Georgios and Choutas, Vasileios and Ghorbani, Nima and Bolkart, Timo and Osman, Ahmed AA and Tzionas, Dimitrios and Black, Michael J},
  booktitle={Proceedings of the IEEE/CVF conference on computer vision and pattern recognition},
  pages={10975--10985},
  year={2019}
}

@article{schulman2017proximal,
  title={Proximal policy optimization algorithms},
  author={Schulman, John and Wolski, Filip and Dhariwal, Prafulla and Radford, Alec and Klimov, Oleg},
  journal={arXiv preprint arXiv:1707.06347},
  year={2017}
}

@article{mcgeer1990passive,
  title={Passive dynamic walking},
  author={McGeer, Tad},
  journal={The international journal of robotics research},
  volume={9},
  number={2},
  pages={62--82},
  year={1990},
  publisher={Sage Publications Sage CA: Thousand Oaks, CA}
}

@INPROCEEDINGS{11203150,
  author={Vu, Cong-Thanh and Lai, Chi-Cheng and Liu, Yen-Chen},
  booktitle={2025 IEEE-RAS 24th International Conference on Humanoid Robots (Humanoids)}, 
  title={Learning Virtual Passive Dynamic Walking Using the Kneed Walker Model for Guiding Policies}, 
  year={2025},
  volume={},
  number={},
  pages={1094-1100},
  doi={10.1109/Humanoids65713.2025.11203150}}

@INPROCEEDINGS{4376312,
  author={Kim, Joohyung and Choi, Chong-Ho and Spong, Mark W.},
  booktitle={2007 IEEE International Conference on Control and Automation}, 
  title={Passive Dynamic Walking with Symmetric Fixed Flat Feet}, 
  year={2007},
  volume={},
  number={},
  pages={24-30},
  doi={10.1109/ICCA.2007.4376312}}

@article{uthai2025opportunities,
  title={Opportunities challenges and roadmap for humanoid robots in construction},
  author={Uthai, Thanakon and You, Hengxu and Wang, Mengjun and Smith, Kaleb and Spackman, Everett and Ryan, Zoe and Li, Shuai and Du, Jing},
  journal={Scientific Reports},
  year={2025},
  publisher={Nature Publishing Group UK London}
}

@article{gu2025humanoid,
  title={Humanoid locomotion and manipulation: Current progress and challenges in control, planning, and learning},
  author={Gu, Zhaoyuan and Li, Junheng and Shen, Wenlan and Yu, Wenhao and Xie, Zhaoming and McCrory, Stephen and Cheng, Xianyi and Shamsah, Abdulaziz and Griffin, Robert and Liu, C Karen and others},
  journal={arXiv preprint arXiv:2501.02116},
  year={2025}
}

@INPROCEEDINGS{9867274,
  author={Iwatani, Yasushi and Kinugasa, Tetsuya},
  booktitle={2022 American Control Conference (ACC)}, 
  title={A necessary condition for passive dynamic walking}, 
  year={2022},
  volume={},
  number={},
  pages={1885-1890},
  doi={10.23919/ACC53348.2022.9867274}}

@INPROCEEDINGS{6027078,
  author={Luo, Ren C. and Chen, Chwan Hsen and Pu, Yi Hao and Chang, Jia Rong},
  booktitle={2011 IEEE/ASME International Conference on Advanced Intelligent Mechatronics (AIM)}, 
  title={Towards active actuated natural walking humanoid robot legs}, 
  year={2011},
  volume={},
  number={},
  pages={886-891},
  doi={10.1109/AIM.2011.6027078}}

@article{cheng2024expressive,
  title={Expressive whole-body control for humanoid robots},
  author={Cheng, Xuxin and Ji, Yandong and Chen, Junming and Yang, Ruihan and Yang, Ge and Wang, Xiaolong},
  journal={arXiv preprint arXiv:2402.16796},
  year={2024}
}

@ARTICLE{10415857,
  author={Tong, Yuchuang and Liu, Haotian and Zhang, Zhengtao},
  journal={IEEE/CAA Journal of Automatica Sinica}, 
  title={Advancements in Humanoid Robots: A Comprehensive Review and Future Prospects}, 
  year={2024},
  volume={11},
  number={2},
  pages={301-328},
  doi={10.1109/JAS.2023.124140}}

@article{10.1145/3450626.3459670,
author = {Peng, Xue Bin and Ma, Ze and Abbeel, Pieter and Levine, Sergey and Kanazawa, Angjoo},
title = {AMP: adversarial motion priors for stylized physics-based character control},
year = {2021},
issue_date = {August 2021},
publisher = {Association for Computing Machinery},
address = {New York, NY, USA},
volume = {40},
number = {4},
issn = {0730-0301},
url = {https://doi.org/10.1145/3450626.3459670},
doi = {10.1145/3450626.3459670},
journal = {ACM Trans. Graph.},
month = jul,
articleno = {144},
numpages = {20}
}

@article{10.1145/3197517.3201311,
author = {Peng, Xue Bin and Abbeel, Pieter and Levine, Sergey and van de Panne, Michiel},
title = {DeepMimic: example-guided deep reinforcement learning of physics-based character skills},
year = {2018},
issue_date = {August 2018},
publisher = {Association for Computing Machinery},
address = {New York, NY, USA},
volume = {37},
number = {4},
issn = {0730-0301},
url = {https://doi.org/10.1145/3197517.3201311},
doi = {10.1145/3197517.3201311},
journal = {ACM Trans. Graph.},
month = jul,
articleno = {143},
numpages = {14}
}

@ARTICLE{10870445,
  author={Yan, Yashuai and Mascaro, Esteve Valls and Egle, Tobias and Lee, Dongheui},
  journal={IEEE Robotics \& Automation Magazine}, 
  title={I-CTRL: Imitation to Control Humanoid Robots Through Bounded Residual Reinforcement Learning}, 
  year={2025},
  volume={32},
  number={1},
  pages={59-67},
  doi={10.1109/MRA.2025.3527284}}

@INPROCEEDINGS{11128480,
  author={Watanabe, Ryo and Li, Chenhao and Hutter, Marco},
  booktitle={2025 IEEE International Conference on Robotics and Automation (ICRA)}, 
  title={DFM: Deep Fourier Mimic for Expressive Dance Motion Learning}, 
  year={2025},
  volume={},
  number={},
  pages={9644-9650},
  doi={10.1109/ICRA55743.2025.11128480}}

@article{zhao2025learning,
  title={Learning aggressive animal locomotion skills for quadrupedal robots solely from monocular videos},
  author={Zhao, Liu and Luo, Zeren and Han, Yimin and Zhang, Jiahui and Chen, Yuanhao and Liu, Yunhui and Lu, Peng},
  journal={npj Robotics},
  volume={3},
  number={1},
  pages={32},
  year={2025},
  publisher={Nature Publishing Group UK London}
}

@INPROCEEDINGS{11093565,
  author={Vu, Cong-Thanh and Huang, Hsin-Hui and Liu, Yen-Chen},
  booktitle={2025 10th International Conference on Control and Robotics Engineering (ICCRE)}, 
  title={Autonomous Navigation for Human-Following Robots Based on Optimized Position Tracking}, 
  year={2025},
  volume={},
  number={},
  pages={23-27},
  doi={10.1109/ICCRE65455.2025.11093565}}

@INPROCEEDINGS{11246444,
  author={Vu, Cong-Thanh and Liu, Yen-Chen},
  booktitle={2025 IEEE/RSJ International Conference on Intelligent Robots and Systems (IROS)}, 
  title={Autonomous Adjustment of Tracking Position in Dynamic Environments for Human-Following Robots Using Deep Reinforcement Learning}, 
  year={2025},
  volume={},
  number={},
  pages={16863-16869},
  doi={10.1109/IROS60139.2025.11246444}}

@InProceedings{10.1007/978-981-16-2094-2_49,
author="Nguyen, Anh-Tu
and Nguyen, Van-Truong
and Nguyen, Xuan-Thuan
and Vu, Cong-Thanh",
editor="Tran, Duc-Tan
and Jeon, Gwanggil
and Nguyen, Thi Dieu Linh
and Lu, Joan
and Xuan, Thu-Do",
title="Development of a Multiple-Sensor Navigation System for Autonomous Guided Vehicle Localization",
booktitle="Intelligent Systems and Networks ",
year="2021",
publisher="Springer Singapore",
address="Singapore",
pages="402--410",
isbn="978-981-16-2094-2"
}

\end{document}